\documentclass[sigconf]{acmart}
\AtBeginDocument{%
  }
\usepackage{graphicx}
\usepackage{subcaption}
\usepackage{caption}
\acmConference[ACM MM, 2025]{Make sure to enter the correct conference title from your rights confirmation email}{October 27--31,2025}{Dublin,Ireland}

\renewcommand\footnotetextcopyrightpermission[1]

\begin{document}

\title{Assessing Color Vision Test in Large Vision-language Models}

\author{Hongfei Ye}
\authornote{Both authors contributed equally to this research.}
\affiliation{%
  \institution{University of Chinese Academy of Sciences}
  \city{Hangzhou}
  \country{China}}
\email{yehongfei23@mails.ucas.ac.cn}

\author{Bin Chen}
\authornotemark[1]
\affiliation{%
  \institution{University of Chinese Academy of Sciences}
  \city{Hangzhou}
  \country{China}}
\email{chenbin232@mails.ucas.ac.cn}

\author{Wenxi Liu}
\affiliation{%
  \institution{University of Chinese Academy of Sciences}
  \city{Beijing}
  \country{China}}
\email{liuwenxi23@mails.ucas.ac.cn}

\author{Yu Zhang}
\affiliation{%
  \institution{University of Chinese Academy of Sciences}
  \city{Hangzhou}
  \country{China}}
\email{zhangyu2312@mails.ucas.ac.cn}

\author{Zhao Li}
\affiliation{%
  \institution{Zhejiang Lab}
  \city{Hangzhou}
  \country{China}}
\email{lzjoey@gmail.com}

\author{Dandan Ni}
\affiliation{%
  \institution{Zhejiang University}
  \city{Hangzhou}
  \country{China}}
\email{nidd@zju.edu.cn}

\author{Hongyang Chen}
\authornote{Corresponding author: \href{mailto:dr.h.chen@ieee.org}{dr.h.chen@ieee.org}}
\affiliation{%
  \institution{Zhejiang Lab}
  \city{Hangzhou}
  \country{China}}
\email{dr.h.chen@ieee.org}

\begin{abstract}
With the widespread adoption of large vision-language models, the capacity for color vision in these models is crucial. However, the color vision abilities of large visual-language models have not yet been thoroughly explored. To address this gap, we define a color vision testing task for large vision-language models and construct a dataset \footnote{Anonymous Github Showing some of the data https://anonymous.4open.science/r/color-vision-test-dataset-3BCD} that covers multiple categories of test questions and tasks of varying difficulty levels. Furthermore, we analyze the types of errors made by large vision-language models and propose fine-tuning strategies to enhance their performance in color vision tests.
\end{abstract}
\settopmatter{printacmref=false}
\maketitle

\section{Introduction}
The ability to perceive and interpret colors accurately is crucial for various applications, such as autonomous driving\cite{cui2024survey,xu2024drivegpt4} and pharmaceutical research\cite{bhattacharya2024large,wang2024scientific}, where Large Vision-Language Models (LVLMs) are increasingly being deployed. The color perception capabilities of these models directly impact their safety and reliability in real-world scenarios. Color vision testing\cite{paramei2023color}, which assesses the ability to distinguish different colors, is essential for evaluating these capabilities. Traditional methods, such as the Ishihara Color Vision Test\cite{birch1997efficiency}, have been widely used to diagnose color vision deficiencies in humans. However, the application of these tests to LVLMs remains underexplored.

\begin{figure}
\centering
\includegraphics[width=1.0\linewidth, height=0.2\textheight]{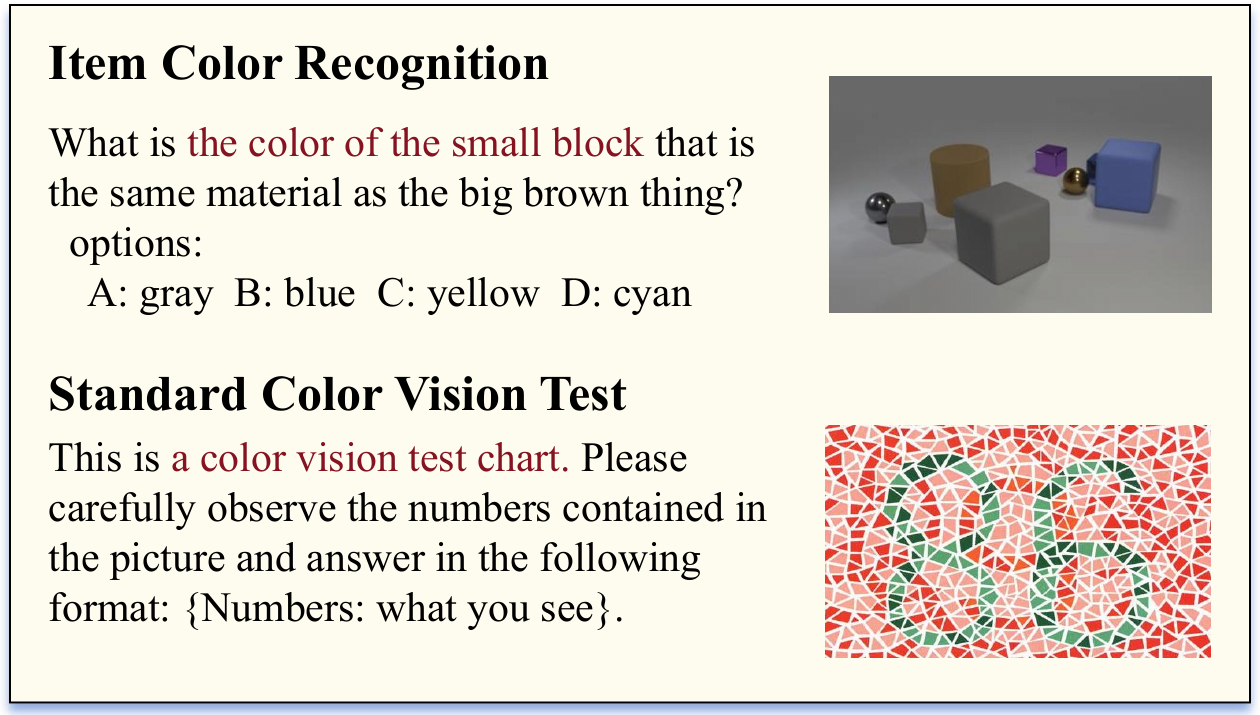} 
\caption{Comparison of previous study and standardized visual test questions }
\label{fig:workcomparision}
\end{figure}

Previous research has primarily focused on evaluating the general capabilities of large vision-language models (LVLMs), such as image captioning and object recognition \cite{liu2024mmbench,yue2024mmmu}. However, these studies \cite{chen2024we,samin2024colorfoil} have not thoroughly explored the models' visual perception abilities, particularly concerning color vision. Specifically, prior investigations \cite{chen2024we} have addressed only a limited range of topics related to color vision within visual language models, representing less than 1\% of the overall dataset. Furthermore, they fail to quantitatively assess the models' ability to discriminate among the three primary colors—red, green, and blue. Additionally, it does not provide a quantitative representation of the severity of color vision deficits in visual language models compared to standard color vision tests, as shown in fig \ref{fig:workcomparision}. This oversight limits our understanding of how these models interpret and process color information, which is critical for their application in safety-sensitive domains. The current research on color vision testing lacks an appropriate dataset and suitable tasks, which has hindered previous studies from thoroughly investigating this area. Our study aims to investigate the boundaries of LVLMs' capabilities in color vision testing. Specifically, we seek to understand how well these models perform when subjected to standardized color vision tests and identify the limitations in their color perception abilities.

We propose a pipeline for evaluating LVLMs on color vision test, which begins with the formalization of these problems by defining their structure and criteria. We then developed two challenging tasks for testing large vision-language models, creating manually verified datasets containing 5450 color vision test images, respectively. Effective evaluation metrics for color vision tests were established, and scoring was performed by multiple evaluators to ensure reliability. The performance of state-of-the-art LVLMs, specifically JanusPro-7B and GPT4o, were assessed, revealing their outstanding capabilities in color vision test. Additionally, we conducted an exploratory analysis of model error types, providing insights that can inform strategies to enhance the color vision test abilities of large vision-language models in future research. To summarize, our contributions are as follows:

\begin{itemize}
    \item We emphasize the significance of evaluating the color perception capabilities of Large Vision-Language Models (LVLMs) and propose a comprehensive set of tasks encompassing two levels of difficulty across five distinct categories.
    \item We develop a specialized dataset tailored for color vision testing, facilitating a systematic evaluation of state-of-the-art (SOTA) models.
    \item We conduct an analysis of the types of errors made by these models and suggest enhancement methods, including LoRA fine-tuning, which notably improve error correction rates.
\end{itemize}

\begin{figure*}[t]
\centering
\includegraphics[width=1.0\textwidth, height=0.33\textheight]{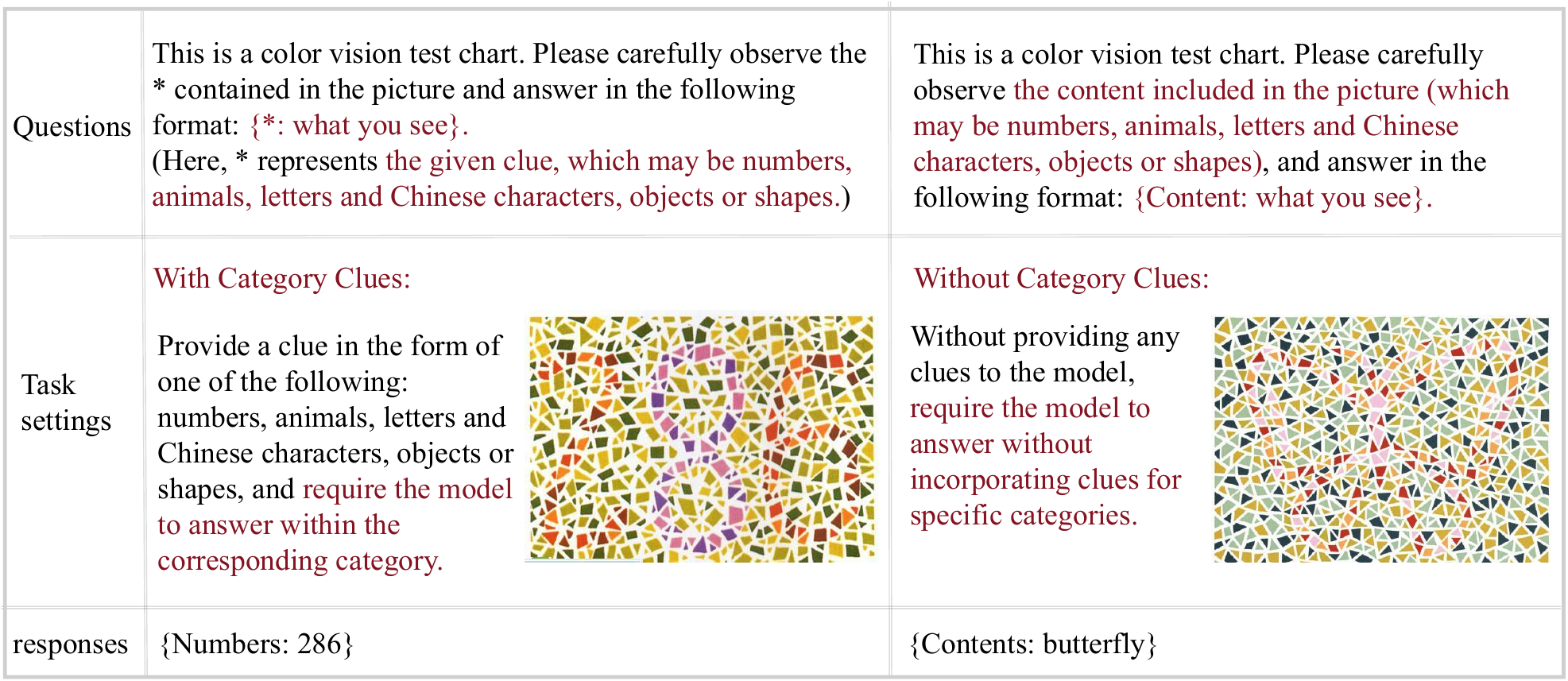} 
\caption{We present examples of our two tasks. The easy task includes cues that indicate the content category of the visual test chart, whereas the hard task does not provide such cues. The answers consist of responses pertaining to the content of the color vision test image.}
\label{fig:dataset}
\end{figure*}

\section{Related Work}
\textbf{Visual Perception}
Visual perception\cite{wade2013visual} refers to the process by which the human brain transforms information patterns from the retina into meaningful interpretations of the world. Low-level visual features\cite{berman2014perception,rouw1997detecting}, such as color, are crucial for human perception, and color vision tests\cite{melamud2004color,stiles1959color} are employed to diagnose color vision deficiencies. The Ishihara test\cite{birch1997efficiency} is one of the most widely used methods for assessing color vision. Previous work\cite{samin2024colorfoil} has explored the color recognition capabilities of traditional multimodal models. However, the abilities of large multimodal models (LVLMs) in tasks such as color vision testing and understanding remain unclear. In contrast to prior research, which overlooked the challenges of color vision for LVLMs, our study specifically investigates their performance in color vision tests.

\textbf{Benchmarks for LVLMs}
Recent advancements in large vision-language models(LVLMs) and visual language models (VLMs) emphasize the need for robust evaluation frameworks. Key benchmarks like MATHVERSE\cite{zhang2024mathverse}, MVP-Bench\cite{li2024mvp}, MMMU\cite{yue2024mmmu}, and SEED-Bench\cite{li2023seed} reveal significant limitations in model performance across various tasks, including visual mathematical reasoning, spatial understanding, and high-level semantic comprehension. Notably, the MMMU Benchmark aims to comprehensively assess the capabilities of large vision-language models (LVLMs) but falls short in exploring visual perception deeply. Additionally, MVP-Bench systematically evaluates LVLMs' multi-level visual perception with both high-level and low-level queries. All these benchmarks do not systematically assess the color perception of large vision-language models.
\begin{figure}
 \includegraphics[width=1.0\columnwidth, height=0.15\textheight]{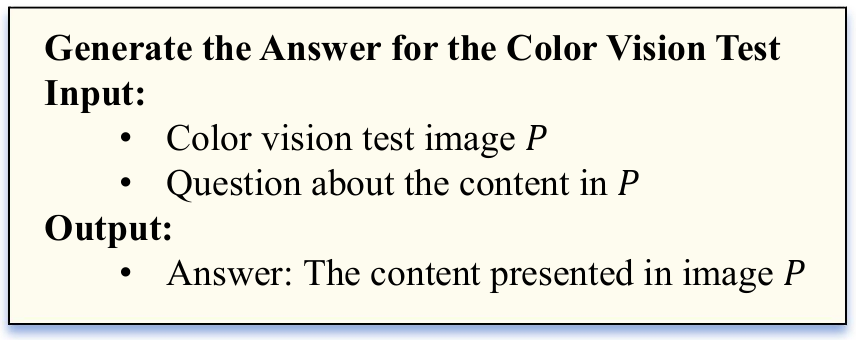}
 \caption{Our setup for the color vision task.}
 \label{fig:colorvisiontest}
\end{figure}

\section{Task Setup and Datasets}
A color vision test can be succinctly described as follows: respondents provide answers based on the content of a given color vision test image. To evaluate the color vision capabilities of large vision-language models (LVLMs), we define our task setup around a color vision test image, denoted as \( P \). The task is structured as shown in Figure \ref{fig:colorvisiontest}. By evaluating the responses, we can gauge the model's abilities in the context of color vision testing. To facilitate this evaluation, we have constructed a dataset that includes a diverse range of color vision test images, each annotated with corresponding answers. This dataset serves as a foundation for rigorous assessment and comparison of various LVLMs regarding their effectiveness in addressing color vision tests.

\subsection{Task Setup}
Our tasks includes two diverse Visual Question Answering (VQA) tasks, each designed with varying levels of difficulty. Color Vision Test Easy (CVTE) encompasses five categories: numbers, animals, letters \& Chinese characters, objects, and shapes, offering a broad spectrum of challenges. Color Vision Test Hard (CVTH), which is more difficult than CVTE, utilizes the same dataset but presents prompts without additional clues about specific categories. As illustrated in Figure \ref{fig:colorvisiontest}, the testing process involves inputting an image and a prompt into the model, which then generates a response. The correctness of the model's response is subsequently evaluated.

\begin{figure*}[t]
\centering
\includegraphics[width=1.0\textwidth, height=0.43\textheight]{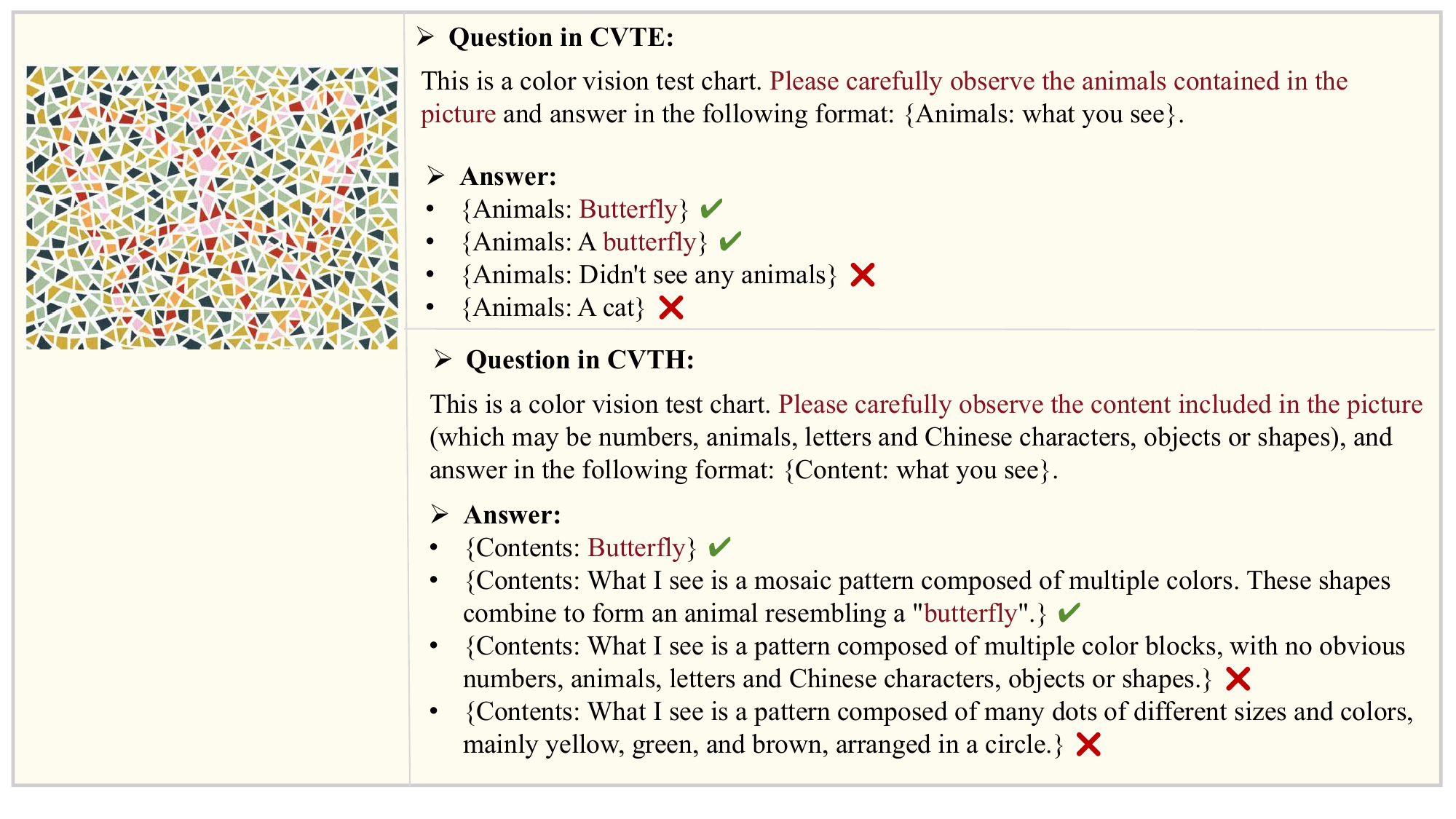} 
\caption{Color Vision Test Setting. Our proposed color vision test encompasses two task settings of varying difficulty. Participants are instructed to carefully observe the animals depicted in the image. In the Color Vision Test for Easy (CVTE), responses should identify the categories of animals present, following the format: \texttt{Animals: (identified category)}. Meanwhile, in the Color Vision Test for Hard (CVTH), participants are required to describe the content within a specified range, utilizing the format: \texttt{What I see is...}.}
\label{fig:evaluation}
\end{figure*}

\subsection{Dataset}
\paragraph{Dataset Design}
The dataset for this benchmark is meticulously designed to rigorously assess the color vision capabilities of multimodal large models. It encompasses a diverse range of images featuring five categories: numbers, animals, letters\&Chinese characters, objects, and shapes, which are commonly utilized in human color vision tests. At the same time, we use red, green, blue and yellow as the main colors and the intermediate colors of red, green, blue and yellow with different saturation to generate data in the data preparation process. The diversity of the data is ensured by a variety of categories and color compositions. To prevent potential data leakage, we generated color vision test images using the Daltonlens toolkit, which is highly regarded in the field of color vision testing and ensures compliance with our required open-source licenses. By collecting pixel maps from publicly available open-source protocols, we employed the Daltonlens toolkit to create diverse color vision test maps featuring various primary colors and saturated color schemes. This approach not only enhances the diversity of our dataset but also safeguards against data leakage during the training process of the LVLM, ultimately ensuring a fair evaluation. The final dataset includes multiple categories, ensuring a broad spectrum of testing scenarios and enhancing the diversity of the evaluation process.

\begin{table}
    \caption{Counts of various data categories in the dataset.}
    \resizebox{0.3\textwidth}{0.075\textheight}{\begin{tabular}{l|r}
        \hline
        \textbf{Category}            & \textbf{Count} \\ \hline
        Numbers                      & 2,680          \\ \hline
        Animals                      & 770            \\ \hline
        Shapes                       & 750            \\ \hline
        Letters or Characters        & 600            \\ \hline
        Objects                      & 650            \\ \hline
    \end{tabular}}
    \label{tab:data_counts}
\end{table}

\paragraph{Quality Assessment}
The initial version of the dataset collected 9,870 entries, with each category containing entries for numbers (4,720), shapes (1,360), animals (1,440), letters or Chinese characters (1,130), and objects (1,220), respectively. We utilize Coblis to assess the quality of color vision test images. Coblis is a widely recognized tool for evaluating the quality of color vision test images in the field. It transforms any picture to simulate how it would appear to individuals with red, green, blue, or complete color blindness. A panel of experts evaluated the data based on predefined criteria, such as the accuracy and consistency of image annotations. Additionally, we compute the Inter-rater Agreement (IRA) using scores from three raters to ensure the reliability of our human evaluations. When responses or proofs fall short of the established quality standards, we initiate a regeneration process. After this rigorous quality assessment, the dataset was divided into easy (Color Vision Test Easy) and hard task (Color Vision Test Hard), incorporating a diverse range of test images across five categories: numbers, animals, letters \& Chinese characters, objects, and shapes. Specifically, the dataset includes 2,680 entries for numbers, 770 for animals, 750 for shapes, 600 for letters \& Chinese characters, and 650 for objects.

\paragraph{Statistical Characteristics}
The dataset exhibits a balanced distribution of different image categories across easy and hard tasks, ensuring a diverse range of challenges. Specifically, it includes 2,680 entries (49.2\%) for numbers, 770 entries (14.1\%) for animals, 750 entries (13.8\%) for shapes, 600 entries ( 11.0\%) for letters or characters, and 650 entries (11.9\%) for objects.

\subsection{Evaluation Metrics}

The evaluation of the model's performance is based on several metrics, including machine metric, model score, and human evaluation score. The machine metric utilize the Meteor score, a commonly used VQA score, which assesses the model's ability to accurately identify and respond to color vision test images. Model score is achieved by inputting the correct answers and the model’s responses into GPT-4, which determines the correctness of the answers. These metrics collectively provide a comprehensive evaluation of the model's color vision capabilities. In addition to machine and model score metrics, we incorporate human evaluation score to ensure a thorough assessment of the generated model responses. Both human raters and GPT-4 use the same scoring criteria to evaluate the accuracy of the model's responses. As shown in fig \ref{fig:evaluation}, when the test image is a butterfly, the model generates descriptions of the content for both CVTE and CVTH. If the model accurately identifies the image content as a butterfly, it receives a score of 1. To guarantee the reliability and confidence of human evaluations, we calculate the Inter-rater Agreement (IRA) based on scores provided by three raters. For scores exhibiting low agreement (less than 0.5), we discard these evaluations and introduce alternative ones. Finally, the average accuracy of the model is calculated for each category (numbers, shapes, animals, letters \& Chinese characters, and objects).

\section{Experiments}

\subsection{Experimental Setup}
Our experiments were conducted on NVIDIA A100 GPUs. The input and output lengths were configured through API parameters to ensure optimal performance. We evaluated our model on three benchmark datasets: MMBench\cite{liu2024mmbench}, MMStar\cite{chen2024we} and our proposed Color Vision Dataset. MM Benchmark evaluates the accuracy of LVLM in perceiving multimodal information such as images and texts. MMStar quantitatively evaluates LVLM's capabilities in terms of coarse (and fine) perception, using Accuracy as the evaluation metric. The evaluation scores for our proposed dataset include machine scores(meteor), model score (acc) and human evaluation score (acc).

\subsection{Models Evaluated}
We tested a range of state-of-the-art models, both closed-source and open-source. The closed-source models evaluated include GPT-4o\cite{hurst2024gpt}, Gemini-2.0-Pro\cite{team2024gemini}, GLM4V-Plus\cite{glm2024chatglm}, Qwen-VL-Max\cite{Qwen-VL}, and GPT4V\cite{2023GPT4VisionSC}, which utilize API access for implementation. Conversely, the open-source models assessed are InternVL2.5-8B\cite{chen2024expanding}, Janus-Pro-7B\cite{chen2025janus}, LLaVANext-7B\cite{liu2024visual}, Qwen2.5VL-7B-Instruct, and Qwen2.5VL-72B-Instruct\cite{qwen2.5-VL}, all of which leverage official open-source weights for their evaluations. These models were selected to provide a comprehensive comparison across different architectures and capabilities.
\subsection{Evaluation Protocol}
The evaluation was performed using a Language Model (LM) evaluator, with GPT-4 serving as the judge. By inputting both the standard answers and the model responses into GPT-4, we enabled it to determine the correctness of the model's answers. The temperature parameter was configured to 0.0001 to ensure deterministic outputs. Each experiment was conducted in triplicate, and the results were averaged to account for variability, thereby providing a more reliable assessment of performance.

\begin{table*}[h]
\caption{Performance Comparison of Various Models Across CVTE and CVTH. Scores for color vision tests computed as the average of three trials. The Met.(Meteor) calculated as the weighted average of meteor in each class. The Avg. calculated as the weighted average of model score in each class. The All calculated as the model score of CVTH. Best results in \textbf{bold} and second best in \underline{underline}.}
\centering
 \resizebox{0.98\textwidth}{!}{\begin{tabular}{@{}l|lllllll|l|l|l|l@{}}
\toprule
                        & \multicolumn{8}{c|}{CVTE}                      & \multicolumn{3}{c}{CVTH}            \\ \midrule
Model        &Met.           & No.         & Anim.           & L\&C & Shp.            & Obj.           & Avg.           & Rank & Met. &All   & Rank\\ \midrule
JanusPro-7B           &\underline{0.4295} & \underline{20.11}        & \textbf{28.70} & \textbf{23.17}     & \underline{23.20}          & \underline{9.85}           & \textbf{20.86} & \textbf{1} &0.3220   & \underline{17.76} & 2    \\
GPT4o             &\textbf{0.4315}     & \textbf{23.81} & 17.40        & \underline{22.33}              & 16.27          & \textbf{11.85} & \underline{20.28}          & 2  &\textbf{0.4163}  & \textbf{18.39} & \textbf{1}    \\
LLaVANext-7B   &0.0054      & 14.22        & 13.64         & 13.50              & \textbf{25.33} & 8.15           & 14.86         & 3 & 0.0006   & 10.83 & 5    \\
Gemini2.0-flash  &0.3893   & 14.40        & \underline{18.31}          & 13.50              & 15.07          & 9.38           & 14.35          & 4  &\underline{0.3511}  & 11.94 & 3    \\
GLM4V-plus   &0.3900         & 17.39        & 12.47          & 10.00              & 12.67          & 6.92           & 13.98         & 5 &0.3082   & 11.03 & 4    \\
QwenVL-max    &0.3636      & 8.69         & 13.64          & 12.00              & 11.07          & 3.85           & 9.50           & 6 &0.3619   & 8.17 & 6    \\
InternVL2.5-8B  &0.3705     & 8.99         & 14.16          & 10.33              & 12.40          & 1.69           & 9.47           & 7 &0.3218   & 5.05 & 8   \\
GPT4V       &0.3281        & 6.72         & 9.74           & 3.67               & 7.47           & 1.69           & 6.31           & 8 &0.3172   & 5.28 & 7   \\
Qwen2.5VL-7B  &0.3212       & 2.84         & 9.74           & 3.50              & 9.47           & 2.15           & 4.72           & 9 &0.3035  & 2.70 & 9   \\
Qwen2.5VL-72B  &0.2583      & 1.64         & 7.79           & 1.00               & 12.40          & 0.00           & 3.72           & 10 &0.2762  & 2.22 & 10   \\ \bottomrule
\end{tabular}}
\label{tab:model_performance}
\end{table*}

\begin{table*}[htbp]
\centering
\caption{Performance Comparison of Various Models Across CVTE and CVTH in various class. The MS(Model Score) calculated accuracy in each class. The Human use the same metrics as the Model Score. Best results in \textbf{bold} and second best in \underline{underline}.}
\resizebox{0.98\textwidth}{!}{
\begin{tabular}{@{}lcc|cc|cc|cc|cc@{}} 
\toprule
\textbf{Model} & \multicolumn{2}{c|}{\textbf{Number}} & \multicolumn{2}{c|}{\textbf{Animal}} & \multicolumn{2}{c|}{\textbf{L\&C}} & \multicolumn{2}{c|}{\textbf{Shape}} & \multicolumn{2}{c}{\textbf{Object}} \\ 
\cmidrule(lr){2-3} \cmidrule(lr){4-5}\cmidrule(lr){6-7}\cmidrule(lr){8-9}\cmidrule(lr){10-11}
& \textbf{MS} & \textbf{Human} & \textbf{MS} & \textbf{Human} & \textbf{MS} & \textbf{Human} & \textbf{MS} & \textbf{Human} & \textbf{MS} & \textbf{Human} \\ 
\midrule

JanusPro-7B & \underline{20.11} & \underline{20.15} & \textbf{28.70} & \textbf{29.87} & \textbf{23.17} & \textbf{23.50} & \underline{23.20} & \underline{24.27} & \underline{9.85} & \underline{11.69} \\
GPT4o      & \textbf{23.81} & \textbf{23.84} & 17.40 & 18.83 & \underline{22.33} & \underline{21.67} & 16.27 & 18.93 & \textbf{11.85} & \textbf{13.69} \\
LLaVANext-7B & 14.22 & 14.22 & 13.64 & 15.19 & 13.50 & 13.83 & \textbf{25.33} & \textbf{24.93} & 8.15 & 9.85 \\
Gemini2.0-flash & 14.40 & 14.48 & \underline{18.31} & \underline{19.35} & 13.50 & 13.33 & 15.07 & 16.13 & 9.38 & 9.23 \\
GLM4V-plus & 17.39 & 17.54 & 12.47 & 12.34 & 10.00 & 10.50 & 12.67 & 14.67 & 6.92 & 8.46 \\
QwenVL-max & 8.69 & 8.69 & 13.64 & 14.55 & 12.00 & 11.83 & 11.07 & 12.13 & 3.85 & 4.62 \\
InternVL2.5-8B & 8.99 & 9.08 & 14.16 & 15.32 & 10.33 & 11.17 & 12.40 & 14.13 & 1.69 & 2.00 \\
GPT4V & 6.72 & 6.75 & 9.74 & 10.13 & 3.67 & 5.00 & 7.47 & 8.27 & 1.69 & 2.31 \\
Qwen2.5VL-7B & 2.84 & 2.95 & 9.74 & 11.69 & 3.50 & 3.17 & 9.47 & 11.07 & 2.15 & 3.23 \\
Qwen2.5VL-72B & 1.64 & 1.72 & 7.79 & 8.57 & 1.00 & 1.67 & 12.40 & 13.73 & 0.00 & 0.00 \\ 
\bottomrule
\end{tabular}
}
\end{table*}

\subsection{Main Result}
\textbf{Q1: Models’ Ability to Understand and Resolve color vision test.} In this section, we evaluate the performance of various models on the CVTE and CVTH. The results are summarized in Table~\ref{tab:model_performance}. For the CVTE, the Janus-Pro-7B model achieved the highest average score of 20.86\%, outperforming GPT-4o, which scored 20.28\%. Notably, the Qwen2.5VL-72B performed significantly worse, with an average score of only 3.72\%. This discrepancy highlights the varying capabilities of models in handling color vision tasks. For the CVTH, the GPT-4o model achieved the highest average score of 18.39\%, outperforming Janus-Pro-7B, which scored 17.76\%. These results suggests that these results highlight the differing strengths of each model based on the presence or absence of category prompts, indicating that JanusPro-7B may be better suited for tasks requiring additional context, while GPT-4o excels in more straightforward assessments.

\textbf{Q2: What is the Performance of LVLMs Handle Color Vision Test across Different Domains?} We further analyzed the performance of the models across different domains. GPT4o demonstrated strong capabilities in number and object recognition, achieving the highest scores in these categories. On the other hand, JanusPro-7B showed superior performance in animal, letter and Chinese, and shape recognition. These results suggest that different models may have specialized strengths in specific sub-domains, which should be considered when selecting a model for particular tasks.

\begin{table}
\caption{The results of model score, human evaluation score, and machine score for the CVTE task. "Ms." refers to the model score rating, while "Human" denotes the human evaluation score. Best results in \textbf{bold} and second best in \underline{underline}.}
\resizebox{0.4\textwidth}{!}{\begin{tabular}{@{}lccc@{}}
\toprule
\textbf{Model} & \textbf{MS.} & \textbf{Human} & \textbf{Meteor} \\ \midrule
JanusPro-7B     & \textbf{20.86} & \textbf{21.45}      & \underline{0.4295} \\
GPT4o           & \underline{20.28}          & \underline{21.01}      & \textbf{0.4315} \\
LLaVANext-7B    & 14.86          & 15.27      & 0.0054 \\
Gemini2.0-flash & 14.35          & 14.64      & 0.3893 \\
GLM4V-plus      & 13.98          & 14.55      & 0.3900 \\
QwenVL-max      & 9.50           & 9.98       & 0.3636 \\
InternVL2.5-8B  & 9.47           & 10.04      & 0.3705 \\
GPT4V           & 6.31           & 6.71       & 0.3281 \\
Qwen2.5VL-7B    & 4.72           & 5.36       & 0.3212 \\
Qwen2.5VL-72B   & 3.72           & 4.13       & 0.2583 \\ \bottomrule
\label{TSE_human_METEOR}
\end{tabular}}
\end{table}

\textbf{Q3: What is the Relationship Between Model Score, Machine Metric, and Human Evaluation Score} We investigated the relationship between model scores, machine metrics (meteor) and human evaluation score, as depicted in table \ref{TSE_human_METEOR} and \ref{fig:Relationship Between TSE, METEOR, and Human Evaluation Scores}. Our analysis revealed that the scores generated by the large language model are more closely aligned with human evaluation score compared to traditional machine metric. This finding underscores the importance of using human-like evaluation criteria when assessing the performance of models on tasks such as color vision examination. The existing machine metrics, such as meteor, may not adequately capture the nuances of human perception, necessitating the use of more sophisticated evaluation frameworks.

\begin{figure}[t]
\centering
\includegraphics[width=1.0\linewidth, height=0.34\textheight]{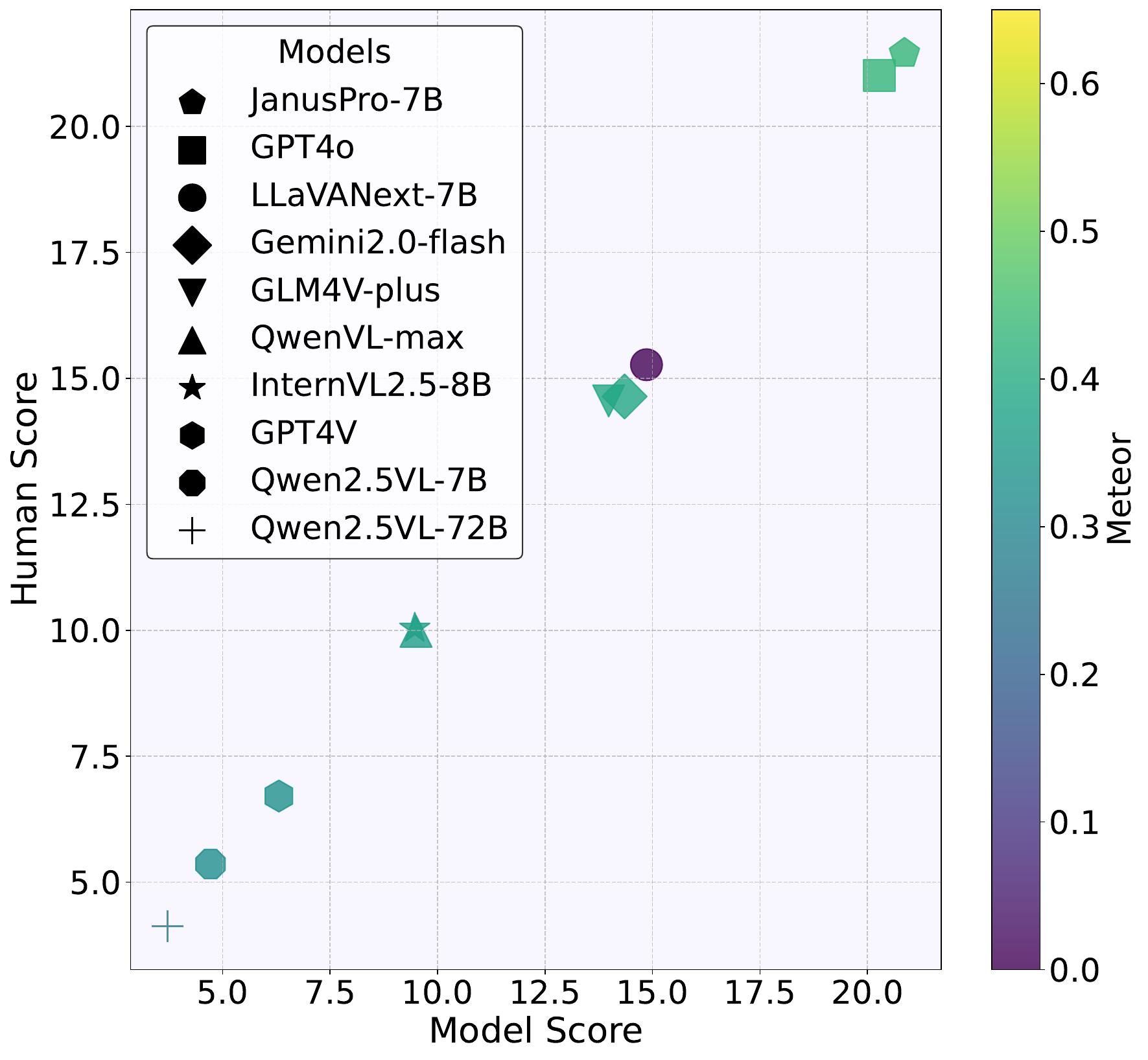} 
\caption{Relationship between Human Evaluation Score, Model Score, and Machine Metric(Meteo) across different models.}
\label{fig:Relationship Between TSE, METEOR, and Human Evaluation Scores}
\end{figure}

\begin{figure} 
    \centering
    \begin{subfigure}[H]{0.434\columnwidth} 
        \centering
        \includegraphics[width=1.0\linewidth]{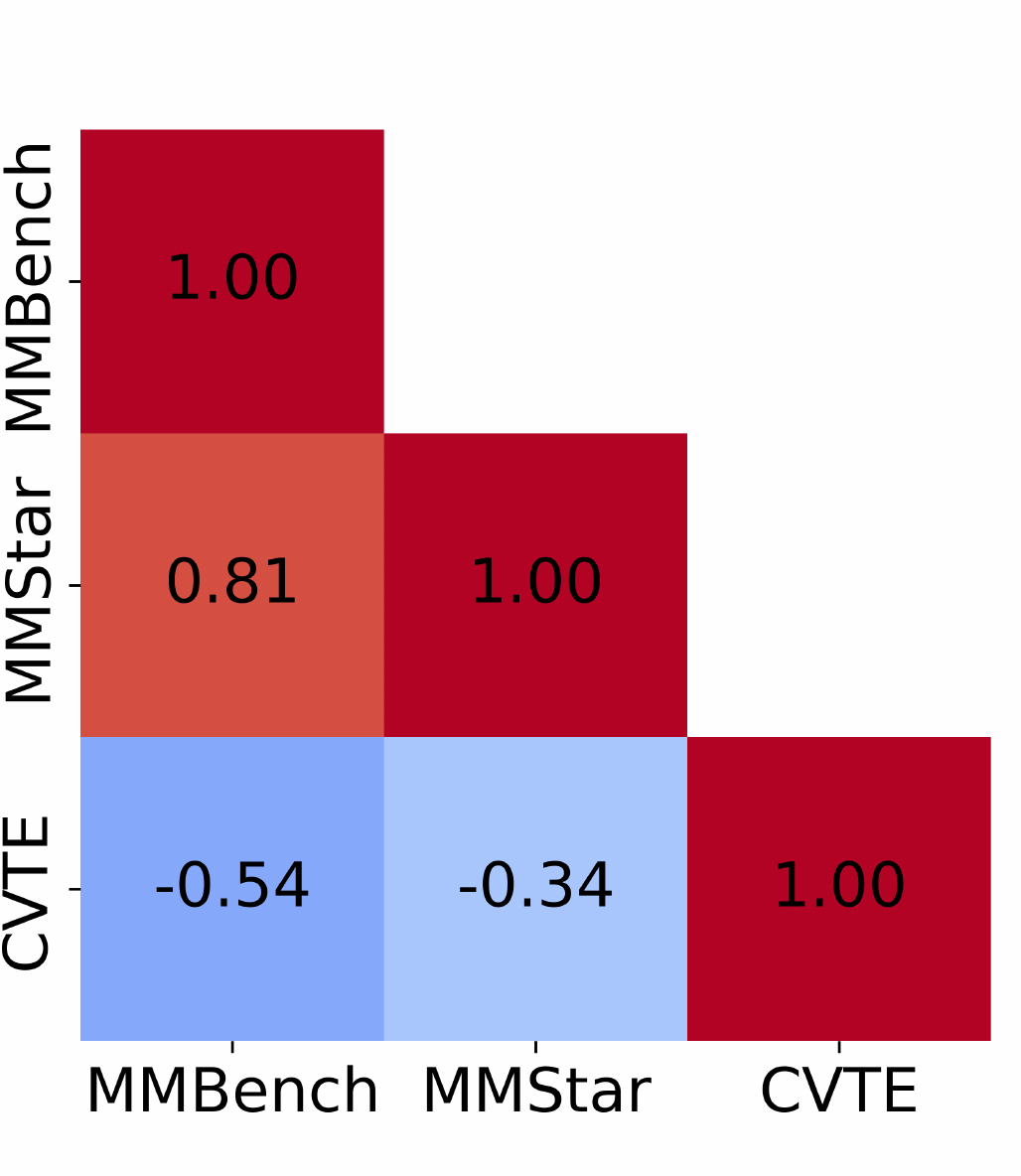} 
        \caption{Easy.}
        \label{fig:correlation_heatmap_easy}
    \end{subfigure}
    \hfill 
    \begin{subfigure}[H]{0.55\columnwidth} 
        \centering
        \includegraphics[width=1.0\linewidth]{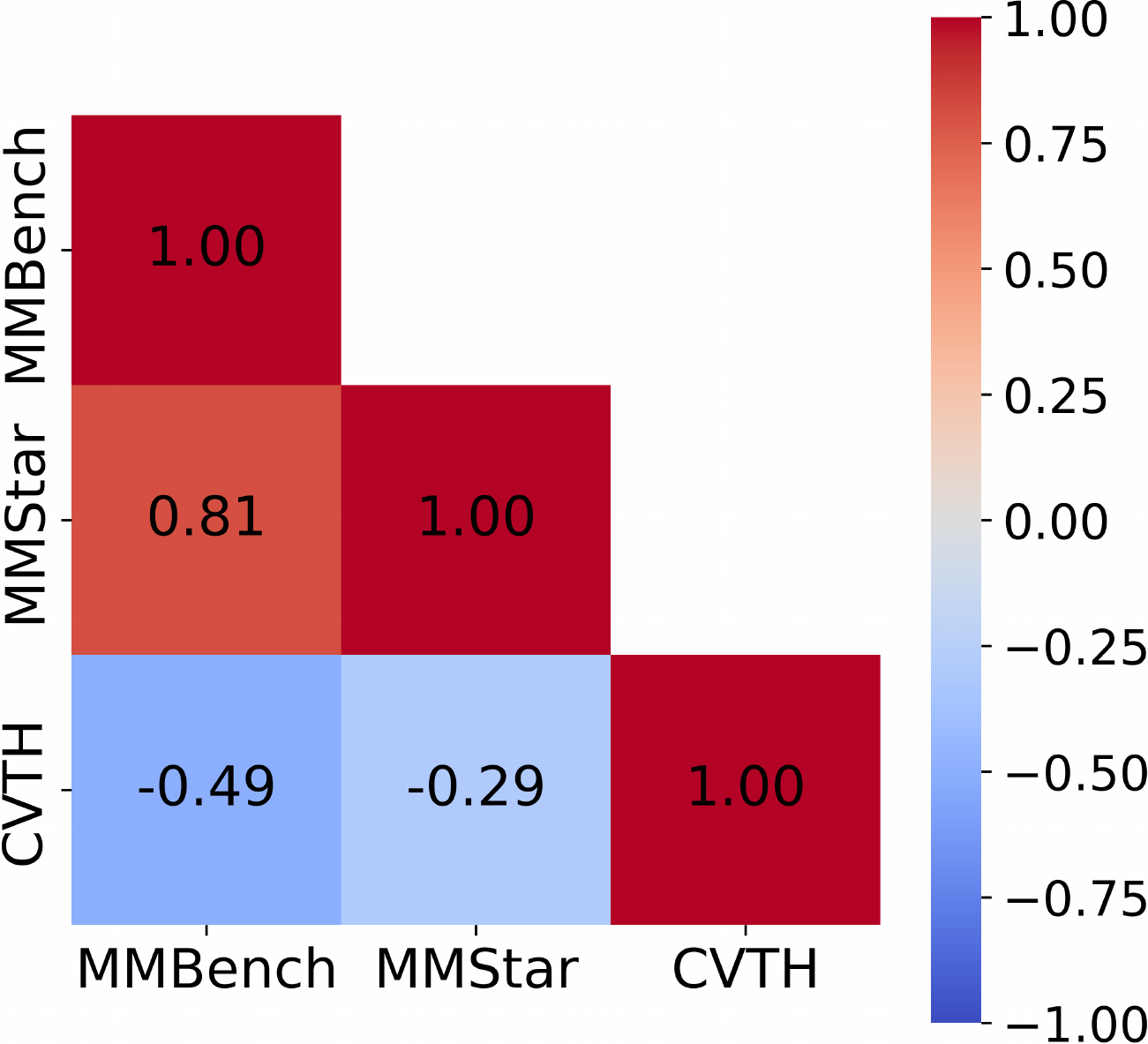} 
        \caption{Hard.}
        \label{fig:correlation_heatmap_hard}
    \end{subfigure}
    \caption{Correlation between other LVLMs benchmarks with color vision tasks.}
    \label{fig:correlation_heatmap}
\end{figure}

\textbf{Q4: What is the Relationship Between the Performance of the Color Vision Test with Other Vision Examination Tasks?} As shown in fig \ref{fig:correlation_heatmap}, we also explored the relationship between the performance of models on color vision examination and other vision examination tasks via Pearson correlation coefficient. Generally, models that performed well on conventional vision tasks  but performed moderately on the color vision task. This indicates that while there is a correlation, the performance on color vision examination is not solely determined by the model's ability to handle general vision tasks.

\subsection{Error Attribution}

\begin{figure}
\centering
\includegraphics[width=0.6\linewidth]{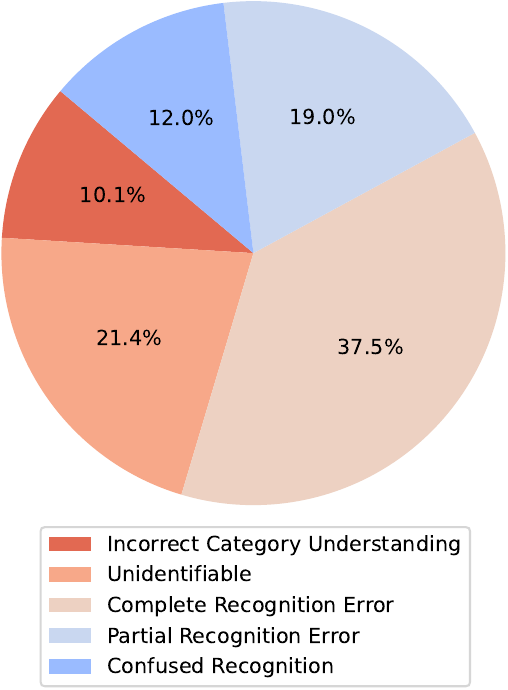} 
\caption{Percentage of response error types}
\label{fig:pie}
\end{figure}

As shown in fig \ref{fig:pie}, the performance of the model across various recognition tasks can be summarized by several error categories: Incorrect Category Understanding (10.14\%) occurs when the model fails to identify the specified category, responding with a different type; Unidentifiable (21.36\%) refers to instances where the model indicates an inability to recognize the content within the image; Complete Recognition Error (37.53\%) occurs when the model's response falls within the same category as the correct answer, yet the response itself is incorrect; Partial Recognition Error (18.98\%) involves the model providing a response that is a part of the correct answer, such as identifying only one of two shapes; and Stochastic Fallback on Uncertainty (11.99\%) indicates that the model consistently produces the same response in the color vision test. For instance, when presented with the digital color vision test input, the model reliably outputs 74.

\begin{figure*}[t]
\centering
\includegraphics[width=1.0\textwidth, height=0.64\textheight]{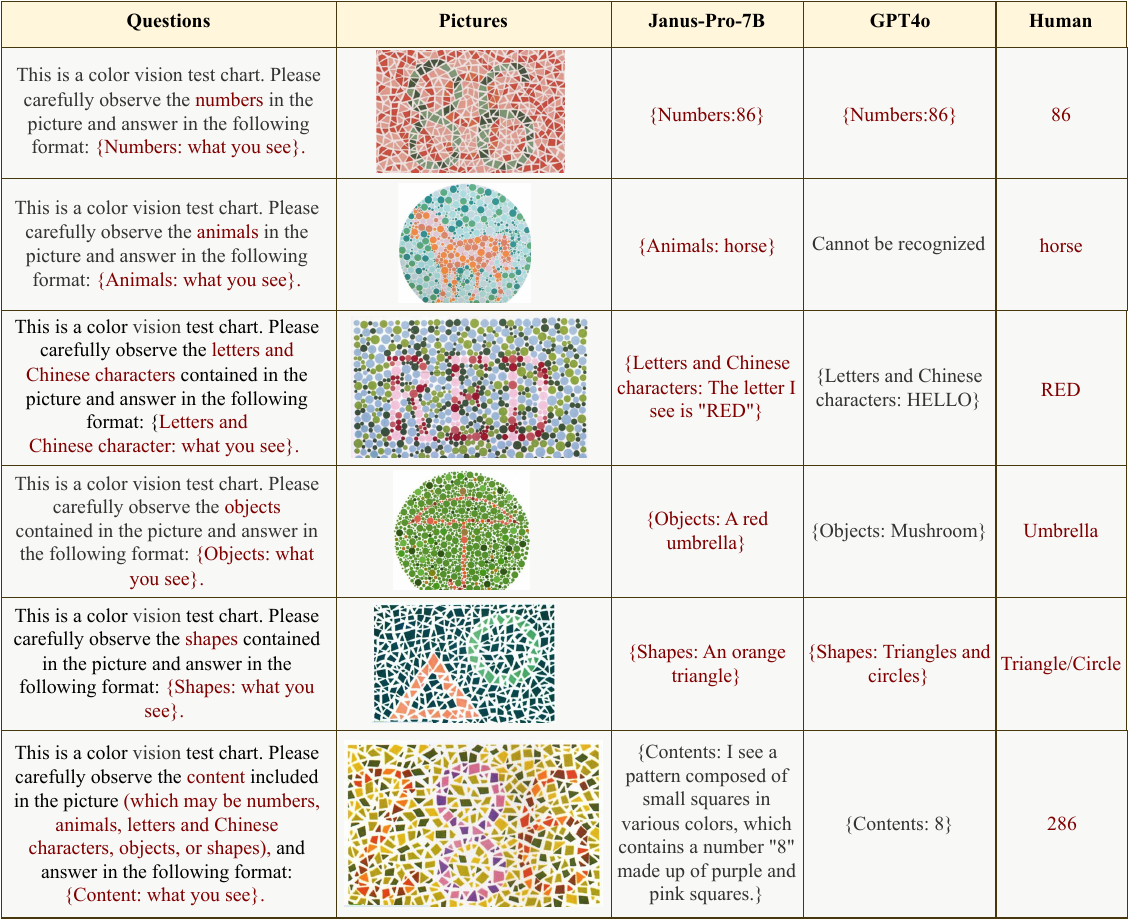} 
\caption{Cases of responses from different LVLMs.}
\label{fig:case study}
\end{figure*}

\subsection{Improvement Experiments}
To address the identified issues, we conducted improvement experiments focusing on model fine-tuning based on LLaVA1.5-7B. We used 70\% of the data for fine-tuning and 30\% for testing. In the CVTE task, the accuracy of LLaVA1.5-7B after LoRA fine-tuning improved from 15.72\% to 94.43\%, representing a significant increase. Similarly, in the CVTH task, the accuracy rose from 11.31\% to 92.23\% after fine-tuning, demonstrating substantial enhancement. Furthermore, the average meteor score for both CVTE and CVTH tasks increased from 0.1924 to 0.9367, indicating a marked improvement in performance. LoRA fine-tuning resulted in a remarkable increase in accuracy, demonstrating the potential for significant performance enhancements through targeted adjustments. However, full fine-tuning led to the model becoming reluctant to provide responses, indicating a need for careful management of fine-tuning strategies to maintain responsiveness while improving performance.

\begin{table}[b]
\caption{Comparison of Performance Before and After LORA Fine-Tuning.}
\begin{tabular}{llll}
\toprule
\textbf{Method}           & \textbf{CVTE}  & \textbf{CVTH}  & \textbf{Met.}    \\ \midrule
LLava1.5-7B+LoRA & 94.43 & 92.23 & 0.9367  \\ 
LLava1.5-7B      & 15.72 & 11.31 & 0.1924 \\ 
\bottomrule
\end{tabular}
\label{tab:performance_comparison}
\end{table}

\subsection{Case Study}
In a case study, we examined a specific question and compared the responses of the top models with a human response. In the simple single-color digit recognition task for "86", both JanusPro-7B and GPT4o performed well. However, in the animal and letter recognition tasks, GPT4o failed to provide correct answers, while JanusPro-7B accurately identified them. For the "umbrella" test image, GPT4o misidentified it as a mushroom, which may reflect the model's data bias. In tasks without additional prompts, GPT4o comprehensively identified the orange triangle and green circle in the image. In the more challenging multi-color digit test image "286", the model could only recognize the digit "8". The human response aligned closely with the best-performing model, further validating the effectiveness of the model's approach.

\section{Conclusion}
The exploration of color vision capabilities in large vision-language models is essential given their widespread adoption. Our proposed color vision testing task, accompanied by a comprehensive dataset, addresses the existing gap in evaluating these models' abilities. Through our analysis of error types, we identify specific areas for improvement and suggest targeted fine-tuning strategies. These enhancements are crucial for advancing the performance of visual language models, ultimately contributing to their effectiveness in tasks requiring nuanced color perception.

\bibliographystyle{ACM-Reference-Format}
\bibliography{main}


\begin{thebibliography}{27}


\ifx \showCODEN    \undefined \def \showCODEN     #1{\unskip}     \fi
\ifx \showISBNx    \undefined \def \showISBNx     #1{\unskip}     \fi
\ifx \showISBNxiii \undefined \def \showISBNxiii  #1{\unskip}     \fi
\ifx \showISSN     \undefined \def \showISSN      #1{\unskip}     \fi
\ifx \showLCCN     \undefined \def \showLCCN      #1{\unskip}     \fi
\ifx \shownote     \undefined \def \shownote      #1{#1}          \fi
\ifx \showarticletitle \undefined \def \showarticletitle #1{#1}   \fi
\ifx \showURL      \undefined \def \showURL       {\relax}        \fi
\providecommand\bibfield[2]{#2}
\providecommand\bibinfo[2]{#2}
\providecommand\natexlab[1]{#1}
\providecommand\showeprint[2][]{arXiv:#2}

\bibitem[Bai et~al\mbox{.}(2023)]%
        {Qwen-VL}
\bibfield{author}{\bibinfo{person}{Jinze Bai}, \bibinfo{person}{Shuai Bai},
  \bibinfo{person}{Shusheng Yang}, \bibinfo{person}{Shijie Wang},
  \bibinfo{person}{Sinan Tan}, \bibinfo{person}{Peng Wang},
  \bibinfo{person}{Junyang Lin}, \bibinfo{person}{Chang Zhou}, {and}
  \bibinfo{person}{Jingren Zhou}.} \bibinfo{year}{2023}\natexlab{}.
\newblock \showarticletitle{Qwen-VL: A Versatile Vision-Language Model for
  Understanding, Localization, Text Reading, and Beyond}.
\newblock \bibinfo{journal}{\emph{arXiv preprint arXiv:2308.12966}}
  (\bibinfo{year}{2023}).
\newblock


\bibitem[Berman et~al\mbox{.}(2014)]%
        {berman2014perception}
\bibfield{author}{\bibinfo{person}{Marc~G Berman}, \bibinfo{person}{Michael~C
  Hout}, \bibinfo{person}{Omid Kardan}, \bibinfo{person}{MaryCarol~R Hunter},
  \bibinfo{person}{Grigori Yourganov}, \bibinfo{person}{John~M Henderson},
  \bibinfo{person}{Taylor Hanayik}, \bibinfo{person}{Hossein Karimi}, {and}
  \bibinfo{person}{John Jonides}.} \bibinfo{year}{2014}\natexlab{}.
\newblock \showarticletitle{The perception of naturalness correlates with
  low-level visual features of environmental scenes}.
\newblock \bibinfo{journal}{\emph{PloS one}} \bibinfo{volume}{9},
  \bibinfo{number}{12} (\bibinfo{year}{2014}), \bibinfo{pages}{e114572}.
\newblock


\bibitem[Bhattacharya et~al\mbox{.}(2024)]%
        {bhattacharya2024large}
\bibfield{author}{\bibinfo{person}{Manojit Bhattacharya},
  \bibinfo{person}{Soumen Pal}, \bibinfo{person}{Srijan Chatterjee},
  \bibinfo{person}{Sang-Soo Lee}, {and} \bibinfo{person}{Chiranjib
  Chakraborty}.} \bibinfo{year}{2024}\natexlab{}.
\newblock \showarticletitle{Large language model to multimodal large language
  model: A journey to shape the biological macromolecules to biological
  sciences and medicine}.
\newblock \bibinfo{journal}{\emph{Molecular Therapy-Nucleic Acids}}
  \bibinfo{volume}{35}, \bibinfo{number}{3} (\bibinfo{year}{2024}).
\newblock


\bibitem[Birch(1997)]%
        {birch1997efficiency}
\bibfield{author}{\bibinfo{person}{Jennifer Birch}.}
  \bibinfo{year}{1997}\natexlab{}.
\newblock \showarticletitle{Efficiency of the Ishihara test for identifying
  red-green colour deficiency}.
\newblock \bibinfo{journal}{\emph{Ophthalmic and Physiological Optics}}
  \bibinfo{volume}{17}, \bibinfo{number}{5} (\bibinfo{year}{1997}),
  \bibinfo{pages}{403--408}.
\newblock


\bibitem[Chen et~al\mbox{.}(2024a)]%
        {chen2024we}
\bibfield{author}{\bibinfo{person}{Lin Chen}, \bibinfo{person}{Jinsong Li},
  \bibinfo{person}{Xiaoyi Dong}, \bibinfo{person}{Pan Zhang},
  \bibinfo{person}{Yuhang Zang}, \bibinfo{person}{Zehui Chen},
  \bibinfo{person}{Haodong Duan}, \bibinfo{person}{Jiaqi Wang},
  \bibinfo{person}{Yu Qiao}, \bibinfo{person}{Dahua Lin}, {et~al\mbox{.}}}
  \bibinfo{year}{2024}\natexlab{a}.
\newblock \showarticletitle{Are We on the Right Way for Evaluating Large
  Vision-Language Models?}
\newblock \bibinfo{journal}{\emph{arXiv preprint arXiv:2403.20330}}
  (\bibinfo{year}{2024}).
\newblock


\bibitem[Chen et~al\mbox{.}(2025)]%
        {chen2025janus}
\bibfield{author}{\bibinfo{person}{Xiaokang Chen}, \bibinfo{person}{Zhiyu Wu},
  \bibinfo{person}{Xingchao Liu}, \bibinfo{person}{Zizheng Pan},
  \bibinfo{person}{Wen Liu}, \bibinfo{person}{Zhenda Xie},
  \bibinfo{person}{Xingkai Yu}, {and} \bibinfo{person}{Chong Ruan}.}
  \bibinfo{year}{2025}\natexlab{}.
\newblock \showarticletitle{Janus-Pro: Unified Multimodal Understanding and
  Generation with Data and Model Scaling}.
\newblock \bibinfo{journal}{\emph{arXiv preprint arXiv:2501.17811}}
  (\bibinfo{year}{2025}).
\newblock


\bibitem[Chen et~al\mbox{.}(2024b)]%
        {chen2024expanding}
\bibfield{author}{\bibinfo{person}{Zhe Chen}, \bibinfo{person}{Weiyun Wang},
  \bibinfo{person}{Yue Cao}, \bibinfo{person}{Yangzhou Liu},
  \bibinfo{person}{Zhangwei Gao}, \bibinfo{person}{Erfei Cui},
  \bibinfo{person}{Jinguo Zhu}, \bibinfo{person}{Shenglong Ye},
  \bibinfo{person}{Hao Tian}, \bibinfo{person}{Zhaoyang Liu}, {et~al\mbox{.}}}
  \bibinfo{year}{2024}\natexlab{b}.
\newblock \showarticletitle{Expanding performance boundaries of open-source
  multimodal models with model, data, and test-time scaling}.
\newblock \bibinfo{journal}{\emph{arXiv preprint arXiv:2412.05271}}
  (\bibinfo{year}{2024}).
\newblock


\bibitem[Cui et~al\mbox{.}(2024)]%
        {cui2024survey}
\bibfield{author}{\bibinfo{person}{Can Cui}, \bibinfo{person}{Yunsheng Ma},
  \bibinfo{person}{Xu Cao}, \bibinfo{person}{Wenqian Ye}, \bibinfo{person}{Yang
  Zhou}, \bibinfo{person}{Kaizhao Liang}, \bibinfo{person}{Jintai Chen},
  \bibinfo{person}{Juanwu Lu}, \bibinfo{person}{Zichong Yang},
  \bibinfo{person}{Kuei-Da Liao}, {et~al\mbox{.}}}
  \bibinfo{year}{2024}\natexlab{}.
\newblock \showarticletitle{A survey on multimodal large language models for
  autonomous driving}. In \bibinfo{booktitle}{\emph{Proceedings of the IEEE/CVF
  Winter Conference on Applications of Computer Vision}}.
  \bibinfo{pages}{958--979}.
\newblock


\bibitem[GLM et~al\mbox{.}(2024)]%
        {glm2024chatglm}
\bibfield{author}{\bibinfo{person}{Team GLM}, \bibinfo{person}{Aohan Zeng},
  \bibinfo{person}{Bin Xu}, \bibinfo{person}{Bowen Wang},
  \bibinfo{person}{Chenhui Zhang}, \bibinfo{person}{Da Yin},
  \bibinfo{person}{Dan Zhang}, \bibinfo{person}{Diego Rojas},
  \bibinfo{person}{Guanyu Feng}, \bibinfo{person}{Hanlin Zhao},
  {et~al\mbox{.}}} \bibinfo{year}{2024}\natexlab{}.
\newblock \showarticletitle{Chatglm: A family of large language models from
  glm-130b to glm-4 all tools}.
\newblock \bibinfo{journal}{\emph{arXiv preprint arXiv:2406.12793}}
  (\bibinfo{year}{2024}).
\newblock


\bibitem[Hurst et~al\mbox{.}(2024)]%
        {hurst2024gpt}
\bibfield{author}{\bibinfo{person}{Aaron Hurst}, \bibinfo{person}{Adam Lerer},
  \bibinfo{person}{Adam~P Goucher}, \bibinfo{person}{Adam Perelman},
  \bibinfo{person}{Aditya Ramesh}, \bibinfo{person}{Aidan Clark},
  \bibinfo{person}{AJ Ostrow}, \bibinfo{person}{Akila Welihinda},
  \bibinfo{person}{Alan Hayes}, \bibinfo{person}{Alec Radford},
  {et~al\mbox{.}}} \bibinfo{year}{2024}\natexlab{}.
\newblock \showarticletitle{Gpt-4o system card}.
\newblock \bibinfo{journal}{\emph{arXiv preprint arXiv:2410.21276}}
  (\bibinfo{year}{2024}).
\newblock


\bibitem[Li et~al\mbox{.}(2023)]%
        {li2023seed}
\bibfield{author}{\bibinfo{person}{Bohao Li}, \bibinfo{person}{Rui Wang},
  \bibinfo{person}{Guangzhi Wang}, \bibinfo{person}{Yuying Ge},
  \bibinfo{person}{Yixiao Ge}, {and} \bibinfo{person}{Ying Shan}.}
  \bibinfo{year}{2023}\natexlab{}.
\newblock \showarticletitle{Seed-bench: Benchmarking multimodal llms with
  generative comprehension}.
\newblock \bibinfo{journal}{\emph{arXiv preprint arXiv:2307.16125}}
  (\bibinfo{year}{2023}).
\newblock


\bibitem[Li et~al\mbox{.}(2024)]%
        {li2024mvp}
\bibfield{author}{\bibinfo{person}{Guanzhen Li}, \bibinfo{person}{Yuxi Xie},
  {and} \bibinfo{person}{Min-Yen Kan}.} \bibinfo{year}{2024}\natexlab{}.
\newblock \showarticletitle{MVP-Bench: Can Large Vision--Language Models
  Conduct Multi-level Visual Perception Like Humans?}
\newblock \bibinfo{journal}{\emph{arXiv preprint arXiv:2410.04345}}
  (\bibinfo{year}{2024}).
\newblock


\bibitem[Liu et~al\mbox{.}(2024b)]%
        {liu2024visual}
\bibfield{author}{\bibinfo{person}{Haotian Liu}, \bibinfo{person}{Chunyuan Li},
  \bibinfo{person}{Qingyang Wu}, {and} \bibinfo{person}{Yong~Jae Lee}.}
  \bibinfo{year}{2024}\natexlab{b}.
\newblock \showarticletitle{Visual instruction tuning}.
\newblock \bibinfo{journal}{\emph{Advances in neural information processing
  systems}}  \bibinfo{volume}{36} (\bibinfo{year}{2024}).
\newblock


\bibitem[Liu et~al\mbox{.}(2024a)]%
        {liu2024mmbench}
\bibfield{author}{\bibinfo{person}{Yuan Liu}, \bibinfo{person}{Haodong Duan},
  \bibinfo{person}{Yuanhan Zhang}, \bibinfo{person}{Bo Li},
  \bibinfo{person}{Songyang Zhang}, \bibinfo{person}{Wangbo Zhao},
  \bibinfo{person}{Yike Yuan}, \bibinfo{person}{Jiaqi Wang},
  \bibinfo{person}{Conghui He}, \bibinfo{person}{Ziwei Liu}, {et~al\mbox{.}}}
  \bibinfo{year}{2024}\natexlab{a}.
\newblock \showarticletitle{Mmbench: Is your multi-modal model an all-around
  player?}. In \bibinfo{booktitle}{\emph{European conference on computer
  vision}}. Springer, \bibinfo{pages}{216--233}.
\newblock


\bibitem[Melamud et~al\mbox{.}(2004)]%
        {melamud2004color}
\bibfield{author}{\bibinfo{person}{Alex Melamud}, \bibinfo{person}{Stephanie
  Hagstrom}, {and} \bibinfo{person}{Elias Traboulsi}.}
  \bibinfo{year}{2004}\natexlab{}.
\newblock \showarticletitle{Color vision testing}.
\newblock \bibinfo{journal}{\emph{Ophthalmic Genetics}} \bibinfo{volume}{25},
  \bibinfo{number}{3} (\bibinfo{year}{2004}), \bibinfo{pages}{159--187}.
\newblock


\bibitem[OpenAI(2023)]%
        {2023GPT4VisionSC}
\bibfield{author}{\bibinfo{person}{OpenAI}.} \bibinfo{year}{2023}\natexlab{}.
\newblock \showarticletitle{GPT-4V(ision) System Card}.
\newblock
\urldef\tempurl%
\url{https://api.semanticscholar.org/CorpusID:263218031}
\showURL{%
\tempurl}


\bibitem[Paramei and Bimler(2023)]%
        {paramei2023color}
\bibfield{author}{\bibinfo{person}{Galina~V Paramei} {and}
  \bibinfo{person}{David~L Bimler}.} \bibinfo{year}{2023}\natexlab{}.
\newblock \showarticletitle{Color vision testing}.
\newblock In \bibinfo{booktitle}{\emph{Encyclopedia of Color Science and
  Technology}}. \bibinfo{publisher}{Springer}, \bibinfo{pages}{517--523}.
\newblock


\bibitem[Rouw et~al\mbox{.}(1997)]%
        {rouw1997detecting}
\bibfield{author}{\bibinfo{person}{Romke Rouw}, \bibinfo{person}{Stephen~M
  Kosslyn}, {and} \bibinfo{person}{Ronald Hamel}.}
  \bibinfo{year}{1997}\natexlab{}.
\newblock \showarticletitle{Detecting high-level and low-level properties in
  visual images and visual percepts}.
\newblock \bibinfo{journal}{\emph{Cognition}} \bibinfo{volume}{63},
  \bibinfo{number}{2} (\bibinfo{year}{1997}), \bibinfo{pages}{209--226}.
\newblock


\bibitem[Samin et~al\mbox{.}(2024)]%
        {samin2024colorfoil}
\bibfield{author}{\bibinfo{person}{Ahnaf~Mozib Samin}, \bibinfo{person}{M~Firoz
  Ahmed}, {and} \bibinfo{person}{Md~Mushtaq~Shahriyar Rafee}.}
  \bibinfo{year}{2024}\natexlab{}.
\newblock \showarticletitle{ColorFoil: Investigating Color Blindness in Large
  Vision and Language Models}.
\newblock \bibinfo{journal}{\emph{arXiv preprint arXiv:2405.11685}}
  (\bibinfo{year}{2024}).
\newblock


\bibitem[Stiles(1959)]%
        {stiles1959color}
\bibfield{author}{\bibinfo{person}{WS Stiles}.}
  \bibinfo{year}{1959}\natexlab{}.
\newblock \bibinfo{title}{Color vision: the approach through
  increment-threshold sensitivity}.
\newblock


\bibitem[Team et~al\mbox{.}(2024)]%
        {team2024gemini}
\bibfield{author}{\bibinfo{person}{Gemini Team}, \bibinfo{person}{Petko
  Georgiev}, \bibinfo{person}{Ving~Ian Lei}, \bibinfo{person}{Ryan Burnell},
  \bibinfo{person}{Libin Bai}, \bibinfo{person}{Anmol Gulati},
  \bibinfo{person}{Garrett Tanzer}, \bibinfo{person}{Damien Vincent},
  \bibinfo{person}{Zhufeng Pan}, \bibinfo{person}{Shibo Wang}, {et~al\mbox{.}}}
  \bibinfo{year}{2024}\natexlab{}.
\newblock \showarticletitle{Gemini 1.5: Unlocking multimodal understanding
  across millions of tokens of context}.
\newblock \bibinfo{journal}{\emph{arXiv preprint arXiv:2403.05530}}
  (\bibinfo{year}{2024}).
\newblock


\bibitem[Team(2025)]%
        {qwen2.5-VL}
\bibfield{author}{\bibinfo{person}{Qwen Team}.}
  \bibinfo{year}{2025}\natexlab{}.
\newblock \bibinfo{title}{Qwen2.5-VL}.
\newblock
\urldef\tempurl%
\url{https://qwenlm.github.io/blog/qwen2.5-vl/}
\showURL{%
\tempurl}


\bibitem[Wade and Swanston(2013)]%
        {wade2013visual}
\bibfield{author}{\bibinfo{person}{Nicholas Wade} {and} \bibinfo{person}{Mike
  Swanston}.} \bibinfo{year}{2013}\natexlab{}.
\newblock \bibinfo{booktitle}{\emph{Visual perception: An introduction}}.
\newblock \bibinfo{publisher}{Psychology Press}.
\newblock


\bibitem[Wang et~al\mbox{.}(2024)]%
        {wang2024scientific}
\bibfield{author}{\bibinfo{person}{Jinge Wang}, \bibinfo{person}{Qing Ye},
  \bibinfo{person}{Li Liu}, \bibinfo{person}{Nancy~Lan Guo}, {and}
  \bibinfo{person}{Gangqing Hu}.} \bibinfo{year}{2024}\natexlab{}.
\newblock \showarticletitle{Scientific figures interpreted by ChatGPT:
  strengths in plot recognition and limits in color perception}.
\newblock \bibinfo{journal}{\emph{NPJ Precision Oncology}} \bibinfo{volume}{8},
  \bibinfo{number}{1} (\bibinfo{year}{2024}), \bibinfo{pages}{84}.
\newblock


\bibitem[Xu et~al\mbox{.}(2024)]%
        {xu2024drivegpt4}
\bibfield{author}{\bibinfo{person}{Zhenhua Xu}, \bibinfo{person}{Yujia Zhang},
  \bibinfo{person}{Enze Xie}, \bibinfo{person}{Zhen Zhao},
  \bibinfo{person}{Yong Guo}, \bibinfo{person}{Kwan-Yee~K Wong},
  \bibinfo{person}{Zhenguo Li}, {and} \bibinfo{person}{Hengshuang Zhao}.}
  \bibinfo{year}{2024}\natexlab{}.
\newblock \showarticletitle{Drivegpt4: Interpretable end-to-end autonomous
  driving via large language model}.
\newblock \bibinfo{journal}{\emph{IEEE Robotics and Automation Letters}}
  (\bibinfo{year}{2024}).
\newblock


\bibitem[Yue et~al\mbox{.}(2024)]%
        {yue2024mmmu}
\bibfield{author}{\bibinfo{person}{Xiang Yue}, \bibinfo{person}{Yuansheng Ni},
  \bibinfo{person}{Kai Zhang}, \bibinfo{person}{Tianyu Zheng},
  \bibinfo{person}{Ruoqi Liu}, \bibinfo{person}{Ge Zhang},
  \bibinfo{person}{Samuel Stevens}, \bibinfo{person}{Dongfu Jiang},
  \bibinfo{person}{Weiming Ren}, \bibinfo{person}{Yuxuan Sun}, {et~al\mbox{.}}}
  \bibinfo{year}{2024}\natexlab{}.
\newblock \showarticletitle{Mmmu: A massive multi-discipline multimodal
  understanding and reasoning benchmark for expert agi}. In
  \bibinfo{booktitle}{\emph{Proceedings of the IEEE/CVF Conference on Computer
  Vision and Pattern Recognition}}. \bibinfo{pages}{9556--9567}.
\newblock


\bibitem[Zhang et~al\mbox{.}(2024)]%
        {zhang2024mathverse}
\bibfield{author}{\bibinfo{person}{Renrui Zhang}, \bibinfo{person}{Dongzhi
  Jiang}, \bibinfo{person}{Yichi Zhang}, \bibinfo{person}{Haokun Lin},
  \bibinfo{person}{Ziyu Guo}, \bibinfo{person}{Pengshuo Qiu},
  \bibinfo{person}{Aojun Zhou}, \bibinfo{person}{Pan Lu},
  \bibinfo{person}{Kai-Wei Chang}, \bibinfo{person}{Yu Qiao}, {et~al\mbox{.}}}
  \bibinfo{year}{2024}\natexlab{}.
\newblock \showarticletitle{Mathverse: Does your multi-modal llm truly see the
  diagrams in visual math problems?}. In \bibinfo{booktitle}{\emph{European
  Conference on Computer Vision}}. Springer, \bibinfo{pages}{169--186}.
\newblock


\end{thebibliography}

\end{document}